\documentclass[11pt]{article}

\usepackage[preprint]{acl}
\usepackage{times}
\usepackage{latexsym}
\usepackage[T1]{fontenc}
\usepackage[utf8]{inputenc}
\usepackage{microtype}
\usepackage{inconsolata}

\usepackage{graphicx}
\usepackage{amsmath, amssymb, amsthm}

\usepackage{hyperref}
\usepackage{url}
\usepackage{stfloats}
\usepackage{natbib}
\usepackage{xcolor}
\usepackage{hyperref}
\usepackage{subcaption}
\usepackage{graphicx}
\usepackage{multirow}
\usepackage{booktabs}
\usepackage{colortbl}
\usepackage{float}
\usepackage{booktabs} 
\usepackage{caption}

\definecolor{dodgerblue}{RGB}{30, 144, 255}
\definecolor{cornflowerblue}{RGB}{100, 149, 237}
\definecolor{myred}{RGB}{186, 62, 69}

\usepackage[utf8]{inputenc} 
\usepackage[T1]{fontenc}    
\usepackage{hyperref}       
\usepackage{url}            
\usepackage{booktabs}       
\usepackage{amsfonts}       
\usepackage{nicefrac}       
\usepackage{microtype}      
\usepackage{xcolor}         
\usepackage{amsmath}
\usepackage{enumitem}
\usepackage[ruled,vlined]{algorithm2e}
\usepackage{wrapfig}
\usepackage{pifont}
\usepackage{fontawesome}
\newcommand{\cmark}{\textcolor{green}{\ding{52}}} 
\newcommand{\xmark}{\textcolor{red}{\ding{56}}}   

\usepackage{listings}
\title{Astra: Activation-Space Tail-Eigenvector Low-Rank Adaptation of Large Language Models}

\author{
  \textbf{Kainan Liu}\textsuperscript{\textdagger}, 
  \textbf{Yong Zhang}\textsuperscript{\textdagger},
  \textbf{Ning Cheng}\textsuperscript{\textasteriskcentered},
  \textbf{Yun Zhu}, \\
  \textbf{Yanmeng Wang},
  \textbf{Shaojun Wang},
  \textbf{Jing Xiao},  \\\\
  \text{Ping An Technology (Shenzhen) Co., Ltd., China} \\
  \text{\href{https://github.com/LyoAI/Astra}{\raisebox{-0.2ex}{\faGithub} \texttt{github.com/LyoAI/Astra}}}
}
\usepackage{float}
\begin{document}
\maketitle

\setlength{\footnotesep}{0.4\baselineskip} 
\renewcommand{\thefootnote}{}\setcounter{footnote}{0}
\footnotemark\footnotetext{%
    \textdagger\ Equal contribution. 
    \textasteriskcentered\ Corresponding author. \\
}
\setcounter{footnote}{0}

\begin{abstract}
Parameter-Efficient Fine-Tuning (PEFT) methods, especially LoRA, are widely used for adapting pre-trained models to downstream tasks due to their computational and storage efficiency. However, in the context of LoRA and its variants, the potential of activation subspaces corresponding to tail eigenvectors remains substantially under-exploited, which may lead to suboptimal fine-tuning performance. In this work, we propose \textbf{Astra} (\underline{A}ctivation-\underline{S}pace \underline{T}ail-Eigenvector Low-\underline{R}ank \underline{A}daptation), a novel PEFT method that leverages the tail eigenvectors of the model output activations—estimated from a small task-specific calibration set—to construct task-adaptive low-rank adapters. By constraining updates to the subspace spanned by these tail eigenvectors, Astra achieves faster convergence and improved downstream performance with a significantly reduced parameter budget. Extensive experiments across natural language understanding (NLU) and natural language generation (NLG) tasks demonstrate that Astra consistently outperforms existing PEFT baselines across 16 benchmarks and even surpasses full fine-tuning (FFT) in certain scenarios.
\end{abstract}

\section{Introduction}
\label{sec:introduction}
Large Language Models (LLMs) have achieved remarkable success across a wide range of tasks~(\citealp{achiam2023gpt}; \citealp{dubey2024llama}; \citealp{guo2025deepseek}). Adapting these pretrained models typically involves full fine-tuning (FFT). Despite its efficacy, FFT incurs substantial computational and memory overhead, limiting its applicability in resource-constrained settings~(\citealp{singh2024study}; \citealp{liu2024grasp}). To overcome these limitations, parameter-efficient fine-tuning (PEFT) methods have been proposed as an effective alternative. By introducing a small number of trainable parameters while keeping the pretrained backbone frozen, PEFT methods substantially reduce training costs while retaining competitive performance~(\citealp{liu2021p}; \citealp{li2021prefix}; \citealp{hu2023llm}).

Among existing PEFT approaches, Low-Rank Adaptation (LoRA)~(\citealp{hu2022lora}) stands out for its simplicity and strong empirical performance. However, the standard initialization of LoRA often yields extremely small gradients at the early training stage, which can hinder optimization and lead to slow convergence or suboptimal adaptation~(\citealp{meng2024pissa}; \citealp{wang2024loraga}). To address this issue, recent studies have explored improved initialization strategies, which can be broadly classified into two categories: \textbf{weight-driven} methods~(\citealp{meng2024pissa}; \citealp{wang2024milora}) that exploit the structure of pretrained weights, and \textbf{data-driven} methods~(\citealp{yang2024corda}; \citealp{wang2024loraga}; \citealp{paischer2024one}) that leverage data distributions or task-specific signals to guide the initialization of low-rank adapters.

However, most existing works overlook an important characteristic of LLM representations: dimensions corresponding to tail eigenvalues remain largely under-utilized during fine-tuning~(\citealp{nayak2025sculpting}). While these directions contribute little to the dominant pretrained representations, they offer substantial flexibility for task-specific adaptation. Adapting model parameters within such under-explored subspaces can effectively increase the model’s representational rank~(\citealp{roy2007effective}), thereby enhancing expressive capacity and improving adaptability to downstream tasks.

Building on this insight, we propose \textbf{Astra} (\underline{A}ctivation-\underline{S}pace \underline{T}ail-Eigenvector Low-\underline{R}ank \underline{A}daptation), a novel PEFT method that exploits under-explored tail subspaces of output activations to construct task-adaptive low-rank adapters. Specifically, Astra first performs eigendecomposition on the covariance matrix of output activations estimated from a small task-specific calibration dataset \(D\), i.e., \(Cov(Y) = Q\Lambda Q^\top\), where \(Q\) denotes the eigenvectors and \(\Lambda\) is the diagonal matrix of corresponding eigenvalues. To explicitly constrain optimization within the under-utilized subspaces, Astra projects the pretrained weight matrix \(W\) onto the subspace spanned by the tail eigenvectors, yielding low-rank adapters aligned with the under-utilized activation directions, i.e., \(A = Q_{[:, -r:]}^\top W\) and \(B = Q_{[:, -r:]}\), where \(r\) denotes the LoRA rank.

This initialization strategy offers several advantages. By focusing adaptation on activation dimensions that are weakly optimized during pretraining, Astra increases the effective representational rank while avoiding interference with dominant pretrained features. As a result, it enables more expressive task-specific updates with improved optimization stability and faster convergence.


We conduct extensive experiments across a diverse set of tasks to evaluate the effectiveness of Astra, covering both natural language understanding (NLU) and natural language generation (NLG) benchmarks. Experimental results show that Astra consistently outperforms existing PEFT baselines on 16 benchmarks and even surpasses full fine-tuning (FFT) on several tasks. Our main contributions can be summarized as follows:

\begin{itemize}[leftmargin=10pt, itemindent=5pt]
\item We propose \textbf{Astra}, a novel LoRA initialization method that exploits under-utilized eigenspaces of output activations to enable effective low-rank adaptation.

\item We conduct extensive evaluations on a broad spectrum of NLU and NLG tasks, including general language understanding, mathematical reasoning, code generation, and commonsense reasoning. The results demonstrate that Astra consistently outperforms strong PEFT baselines and exhibits robust adaptability across tasks.

\item We perform comprehensive ablation studies on eigenvector selection, LoRA rank, and calibration data, systematically validating the effectiveness and efficiency of Astra. 

\item We provide an effective-rank analysis that empirically supports our core hypothesis: adapting within under-explored activation subspaces enhances task-specific representational capacity and improves downstream task performance.
\end{itemize}

\section{Related Work}
\label{sec:related_work}

\paragraph{PEFT.} Parameter-efficient fine-tuning (PEFT) provides a lightweight alternative to full fine-tuning by updating only a small subset of parameters, reducing computational overhead while maintaining strong downstream performance. PEFT methods can be broadly categorized into: (i) \textit{prompt-based}~(\citealp{li2021prefix}; \citealp{liu2021p}), which prepend trainable tokens or embeddings to the input; (ii) \textit{adapter-based}~(\citealp{houlsby2019parameter}; \citealp{ruckle2020adapterdrop}; \citealp{hu2023llm}), which insert small trainable modules into each transformer layer; and (iii) \textit{LoRA-based}~(\citealp{hu2022lora}; \citealp{dettmers2023qlora}), which employ low-rank reparameterization to enable efficient adaptation.

\paragraph{LoRA and Its Variants.} Low-rank adaptation (LoRA) has received significant attention for enabling fine-tuning without altering the original architecture or adding inference overhead~(\citealp{li2018measuring}; \citealp{aghajanyan-etal-2021-intrinsic}). Building on LoRA, subsequent research has explored several directions to enhance its flexibility and efficiency:

\textit{\textbf{Dynamic rank allocation}} methods~(\citealp{valipour2022dylora}; \citealp{liu-etal-2024-alora}), such as AdaLoRA~(\citealp{zhang2023adalora}), adaptively distribute parameter budgets across weight matrices based on importance scores. 

\textit{\textbf{Structural modifications}}~(\citealp{liu2024dora}; \citealp{feng2024mixture}; \citealp{li2024mixlora}) generalize LoRA beyond its original design. For example, DoRA~(\citealp{liu2024dora}) decouples learning into magnitude and direction. 

\textit{\textbf{Hyperparameter optimization}} has also been explored to improve fine-tuning efficiency~(\citealp{kalajdzievski2023rank}; \citealp{lora+2024}). For instance, LoRA+~(\citealp{lora+2024}) introduces differential learning rates for the low-rank matrices A and B, with a higher learning rate for B to accelerate convergence.

\textit{\textbf{Initialization Strategies.}} Recent efforts have also explored initialization strategies to stabilize training and accelerate convergence~(\citealp{meng2024pissa}; \citealp{wang2024milora}; \citealp{yang2024corda}; \citealp{wang2024loraga}), which can be divided into \textbf{weight-driven} and \textbf{data-driven} approaches. 

Weight-driven methods primarily analyze the static geometric properties of pre-trained parameters. For example, PiSSA~(\citealp{meng2024pissa}) applies Singular Value Decomposition (SVD) to the pretrained weight matrix \(W^{(0)}\), capturing the principal singular values to initialize the adapters while relegating the residual components to the frozen weights. While effective, weight-driven methods neglect the dynamic input distribution encountered during inference. In contrast, data-driven approaches incorporate task-specific information into the initialization process. LoRA-GA~(\citealp{wang2024loraga}) introduces a novel initialization scheme that aligns the initial gradients of the low-rank update matrices with those of full fine-tuning.

Our proposed framework, Astra, represents a distinct paradigm within the data-driven category. Unlike gradient-based methods like LoRA-GA that require computationally expensive gradient estimation, Astra operates on the activation-space covariance. By identifying the tail subspace of the output activations, Astra ensures that the adapters are initialized within directions that are statistically under-utilized by the pre-trained model yet functionally critical for the downstream task. This unique activation-centric perspective allows Astra to capture task-specific functional redundancies that are overlooked by purely geometric weight analysis during adaptation. A detailed comparison of LoRA variants is provided in Appendix~\ref{appendix:lora_variants_overview}.

\section{Method}
\label{sec:method}

We introduce Astra, a LoRA initialization method based on the observation that pretrained models exhibit uneven utilization across activation dimensions. In standard LoRA, low-rank adapters are initialized without considering this structural property, which may lead to inefficient early-stage optimization. Astra initializes LoRA updates within activation subspaces associated with smaller eigenvalues, aiming to make use of under-utilized directions during task-specific adaptation.
\subsection{Preliminaries of LoRA's Initialization}
LoRA~(\citealp{hu2022lora}) introduces trainable updates by reparameterizing weight modifications as the product of two low-rank matrices. Formally, given a pretrained weight matrix \( W_0 \in \mathbb{R}^{m \times n} \), LoRA expresses the adapted weight as:
\begin{equation}
    \tilde{W} = W_0 + \Delta W = W_0 + \frac{\alpha}{r} BA
\end{equation}
where \( \Delta W \) denotes the weight change, which is decomposed into two low-rank matrices \( B \in \mathbb{R}^{m \times r} \) and \( A \in \mathbb{R}^{r \times n} \) with an intrinsic rank \( r \ll \min(m, n) \), \(\alpha\) is a scaling constant. This parameterization reduces the number of trainable parameters from \( mn \) to \( (m+n)r \), significantly improving fine-tuning efficiency. In practice, \( A \) is initialized from the Gaussian distribution, while \( B \) is initialized as an all-zero matrix to ensure that the initial model output remains unchanged. However, such random initialization can lead to slower convergence, as the gradients of the trainable adapters can be very small or in random directions during the early stages of fine-tuning~(\citealp{meng2024pissa}). 

\subsection{Activation-Space Tail-Eigenvector Low-Rank Adaptation}
Astra initializes LoRA adapters by constraining low-rank updates to a task-specific tail subspace of the output activation space. The core mechanism involves two stages: (1) performing orthogonal decomposition of the output activations and (2) projecting the pre-trained weights onto the tail subspace to initialize the LoRA adapters. The detailed formulation of each step is presented below.

\paragraph{Step 1: Orthogonal Decomposition.} Astra begins by characterizing the structure of the output activations relevant to the downstream task through covariance analysis. Specifically, we construct a calibration dataset $X = \{x_i\}_{i=1}^N$ by randomly sampling $N=64$ samples from the training data. Let $Y = W X + b \in \mathbb{R}^{d_{\text{out}} \times N}$ denote the collective output activations of a linear layer. We then compute the corresponding covariance matrix as:
\begin{equation}
    \mathrm{Cov}(Y) = \mathbb{E}[Y Y^\top] - \mathbb{E}[Y] \, \mathbb{E}[Y]^\top
\end{equation}
where \(\mathbb{E}[\cdot]\) is the expectation operator. Since \(\mathrm{Cov}(Y)\) is positive semi-definite, it admits an eigendecomposition \(\mathrm{Cov}(Y) = Q \Lambda Q^\top\).

Here, the orthogonal matrix $Q = [Q_{\text{main}} \,|\, Q_{\text{tail}}]$ partitions the eigenvectors into two subspaces, where
\(Q_{\text{main}} \in \mathbb{R}^{d_{\text{out}} \times (d_{\text{out}}-r)}\) corresponds to the principal subspace associated with the dominant eigenvalues, and
\(Q_{\text{tail}} \in \mathbb{R}^{d_{\text{out}} \times r}\) spans the residual subspace defined by the tail eigenvalues. The two subspaces are orthogonal and satisfy \(Q_{\text{main}}^\top Q_{\text{tail}} = 0\). Utilizing these orthonormal bases, we perform an orthogonal decomposition of the output activations $Y$ into two mutually orthogonal projections:
\begin{align}
Y
&= Y_{\text{main}} + Y_{\text{tail}} \\[4mm] 
&= Q_{\text{main}} Q_{\text{main}}^\top Y 
+ Q_{\text{tail}} Q_{\text{tail}}^\top Y
\label{eq:decompose2} \\[4mm]
&= \underbrace{Q_{\text{main}} Q_{\text{main}}^\top W x + b}_{\text{Frozen}}
\;+\;
\underbrace{Q_{\text{tail}} Q_{\text{tail}}^\top W x}_{\text{Trainable}}
\label{eq:decompose3}
\end{align}
where \(Y_{\text{main}} := Q_{\text{main}} Q_{\text{main}}^\top Y\)
and \(Y_{\text{tail}} := Q_{\text{tail}} Q_{\text{tail}}^\top Y\). Given \(Q_{\text{main}}^\top Q_{\text{tail}} = 0\), it follows that $Y_{\text{main}}^\top Y_{\text{tail}} = 0$, ensuring that the two components occupy non-interfering subspaces.

This orthogonal decomposition forms the basis for restricting subsequent parameter updates to the residual activation subspace spanned by \(Q_{\text{tail}}\), thereby exploiting directions that are weakly activated under the pretrained model yet remain amenable to task-specific modulation.

\paragraph{Step 2: Tail-Subspace Projection.} Building upon the decomposition in Eq.\ref{eq:decompose3}, Astra initializes the two learnable low-rank matrices \(A \in \mathbb{R}^{r \times d_{\text{in}}}\) and \(B \in \mathbb{R}^{d_{\text{out}} \times r}\) by projecting the pre-trained weights onto the identified tail subspace. Formally, the initialization is defined as:
\begin{equation}
\label{eq:a_init}
A_{\text{init}} = Q_{\text{tail}}^\top W \in \mathbb{R}^{r \times d_{\text{in}}}.
\end{equation}
\begin{equation}
\label{eq:b_init}
B_{\text{init}} = Q_{\text{tail}} \in \mathbb{R}^{d_{\text{out}} \times r}, 
\end{equation}
where \(A_{\text{init}}\) and \(B_{\text{init}}\) serve as the two learnable low-rank matrices in LoRA. Since the resulting update \(\Delta W = BA\) is non-zero at initialization, we adjust the frozen component to ensure that the original model outputs remain unchanged. Formally, this yields:
\begin{equation}
W^{\prime} = W^{(0)} + \Delta W = (\underbrace{W^{(0)} - B_{\text{init}}A_{\text{init}}}_{\text{Frozen}}) + \underbrace{B^{\prime}A^{\prime}}_{\text{Trainable}}
\end{equation}
where the learnable matrices \(A^{\prime}\) and \(B^{\prime}\) parameterize the task-specific update \(\Delta W\). By constraining the optimization to the subspace spanned by the tail eigenvectors, Astra effectively utilizes previously under-explored directions, thereby enhancing adaptation efficiency. The complete procedure is summarized in Algorithm \ref{alg:Astra}, and a PyTorch-style implementation is provided in Appendix \ref{appendix:pesudocode}.

\begin{algorithm}[h]
\caption{Astra: Activation-Space Tail-Eigenvector Low-Rank Adaptation}
\label{alg:Astra}
\KwIn{Model $M$, LoRA rank $r$, calibration data $x$, weight matrices 
      $W \in \mathbb{R}^{d_{\text{out}} \times d_{\text{in}}}$}
\KwOut{Initialized parameters $W_{\text{frozen}}, A_{\text{init}}, B_{\text{init}}$}

1: $\hat{Y} \gets M(x; W)$ \hfill $\triangleright$ Forward propagation \\[2mm]
2: $\mathrm{Cov}(Y) \gets \mathbb{E}[YY^\top] - \mathbb{E}[Y] \, \mathbb{E}[Y]^\top$ \\[2mm]
3: $\mathrm{Cov}(Y) = Q \Lambda Q^\top$ \hfill $\triangleright$ Eigen-decomposition \\[2mm]
4: Initialize trainable low-rank matrices: \\
\quad $A_{\text{init}} = Q_{\text{tail}}^\top W \in \mathbb{R}^{r \times d_{\text{in}}}$ \\
\quad $B_{\text{init}} = Q_{\text{tail}} \in \mathbb{R}^{d_{\text{out}} \times r}$ \hfill $\triangleright$ Astra Initialization \\[2mm]
5: Compute frozen and update terms: \\
\quad $W_{\text{frozen}} = W^{(0)} - B_{\text{init}} A_{\text{init}}$ \\
\quad $W_{\text{trainable}} = B_{\text{init}} A_{\text{init}}$ \\[2mm]
\KwRet{$W_{\text{frozen}}, A_{\text{init}}, B_{\text{init}}$}
\end{algorithm}

\section{Experiments}
\label{sec:experiments}
In this section, we provide a comprehensive evaluation of Astra from three perspectives. 1) We first assess the Natural Language Understanding (NLU) capabilities using the GLUE~(\citealp{wang2018glue}) benchmark (Section~\ref{sec:experiment_nlu}). 2) Next, we evaluate the performance of our method on Natural Language Generation (NLG) tasks, covering mathematical reasoning, code generation, and commonsense reasoning (Section~\ref{sec:experiment_nlg}). 3) Finally, we conduct ablation studies to analyze the effectiveness of our approach with respect to varying eigenvectors, LoRA ranks and calibration datasets (Section~\ref{sec:experiment_ablation}). All experiments are conducted on NVIDIA A100-SXM4 (80GB) GPUs. 

\subsection{Baselines}
\label{sec:baselines}
To substantiate the effectiveness of our method, we compare Astra against full fine-tuning (FFT), vanilla LoRA, and 6 representative LoRA variants. These variants can be grouped as follows:
\vspace{-2mm}
\begin{enumerate}[leftmargin=10pt, itemindent=2pt]
    \item \textbf{Weight-driven initialization variants}:
    \begin{itemize}[leftmargin=10pt, label={--}]
        \item \textit{PiSSA}~(\citealp{meng2024pissa}) initializes adapters with principal components and freezes the residual.
        \item \textit{MiLoRA}~(\citealp{wang2024milora}) initializes adapters with the smallest singular components.
    \end{itemize}
    \item \textbf{Data-driven initialization variants}:
    \begin{itemize}[leftmargin=10pt, label={--}]
        \item \textit{CorDA}~(\citealp{yang2024corda}) builds adapters conditioned on context for task-specific adaptations.
        \item \textit{LoRA-GA}~(\citealp{wang2024loraga}) constructs low-rank matrices by approximating the gradient from the first step of full fine-tuning.
    \end{itemize}
    \item \textbf{Other LoRA variants} (with modified structure, hyperparameters, etc.):
    \begin{itemize}[leftmargin=10pt, label={--}]
        \item \textit{rsLoRA}~(\citealp{kalajdzievski2023rank}) introduces a square-root scaling factor to LoRA.
        \item \textit{DoRA}~(\citealp{liu2024dora}) decomposes pretrained weights into magnitude and direction components, tuning the magnitude and direction matrix separately.
    \end{itemize}
\end{enumerate}

\begin{table*}[htbp]
\centering
\resizebox{1\linewidth}{!}{%
    \begin{tabular}{cccccccc}
    \toprule[1.2pt]
    & \textbf{\#Params} & \begin{tabular}[c]{@{}c@{}}\textbf{MNLI}\\393k\end{tabular} & \begin{tabular}[c]{@{}c@{}}\textbf{SST-2}\\67k\end{tabular} & \begin{tabular}[c]{@{}c@{}}\textbf{QNLI}\\105k\end{tabular} & \begin{tabular}[c] {@{}c@{}}\textbf{CoLA}\\8.5k\end{tabular} & \begin{tabular}[c]{@{}c@{}}\textbf{MRPC}\\3.7K\end{tabular} & \textbf{Average} \\
    \midrule
    Full  FT & 223M & 86.95$_{\pm 0.04}$ & 97.02$_{\pm 0.03}$ & 98.78$_{\pm 0.02}$ & 84.52$_{\pm 0.01}$ & 84.19$_{\pm 0.05}$ & 90.29 \\
    \midrule
    LoRA  & 3.2M  & 86.97$_{\pm 0.01}$ & 96.62$_{\pm 0.02}$ & 98.75$_{\pm 0.03}$ & 49.95$_{\pm 1.33}$ &  47.67$_{\pm 0.06}$ & 75.99 \\
    DoRA  & 3.4M  & 87.05$_{\pm 0.02}$ & \textbf{97.19}$_{\pm 0.01}$ & 98.79$_{\pm 0.02}$ & 84.23$_{\pm 0.03}$ & 49.88$_{\pm 0.05}$ & 83.43 \\
    rsLoRA & 3.2M  & 87.06$_{\pm 0.01}$ & 97.13$_{\pm 0.02}$ & 98.79$_{\pm 0.02}$ & 83.89$_{\pm 0.02}$ & 49.63$_{\pm 0.04}$ & 83.30 \\
    \midrule
    PiSSA & 3.2M  & 87.01$_{\pm 0.01}$ & 97.08$_{\pm 0.01}$ & 98.82$_{\pm 0.01}$ & 84.80$_{\pm 0.01}$ & 82.84$_{\pm 0.01}$ & 90.11 \\
    CorDA & 3.2M  & \textbf{87.11}$_{\pm 0.03}$ & \textbf{97.19}$_{\pm 0.02}$ & 98.81$_{\pm 0.05}$ & 84.71$_{\pm 0.22}$ & 69.12$_{\pm 0.23}$ & 87.39 \\
    LoRA-GA & 3.2M  & 87.07$_{\pm 0.01}$ & 97.13$_{\pm 0.02}$ & \textbf{98.83}$_{\pm 0.01}$ & 84.76$_{\pm 0.11}$ & 84.19$_{\pm 0.14}$ & 90.40 \\
    \midrule
    \cellcolor[rgb]{ .855,  .949,  .816}Ours  & \cellcolor[rgb]{ .855,  .949,  .816}3.2M  & \cellcolor[rgb]{ .855,  .949,  .816}87.09$_{\pm 0.01}$ & \cellcolor[rgb]{ .855,  .949,  .816}96.45$_{\pm 0.01}$ & \cellcolor[rgb]{ .855,  .949,  .816}\textbf{98.83}$_{\pm 0.01}$ & \cellcolor[rgb]{ .855,  .949,  .816}\textbf{87.87}$_{\pm 0.06}$ &  \cellcolor[rgb]{ .855,  .949,  .816}\textbf{88.36}$_{\pm 0.12}$ & \cellcolor[rgb]{ .855,  .949,  .816}\textbf{91.72} \\
    \bottomrule[1.2pt]
    \end{tabular}%
}
\caption{Performance of T5-base fine-tuned with different adaptation methods on 5 datasets of the GLUE benchmark. We report accuracy for all tasks, and the results are averaged over three runs with different random seeds. Bold values indicate the best performance.}
\label{tab:NLU}
\vspace{-2mm}
\end{table*}%

\subsection{Natural Language Understanding}
\label{sec:experiment_nlu}

\paragraph{Models and Datasets.} We fine-tune the T5-base model~\citep{raffel2020exploring} on a subset of tasks from the GLUE benchmark~\citep{wang2018glue}, including MNLI, QNLI, SST-2, CoLA and MRPC. The model is evaluated on the corresponding development sets, and accuracy is reported as the evaluation metric for all tasks. Additional details regarding the benchmarks are presented in Appendix~\ref{appendix:benchmark_nlu}.

\paragraph{Implementation Details.} We follow the experimental setup described in~\citep{wang2024loraga} to ensure a fair comparison. Specifically, we convert the labels into tokens (e.g., "positive" or "negative") and use the prompt tuning to fine-tune the model for 1 epoch on each dataset. The normalized probabilities assigned to these tokens are then used for classification. Further experimental setup and implementation details can be found in Appendix~\ref{appendix:implementation_details_nlu}.

\paragraph{Main Results.} Table~\ref{tab:NLU} presents the performance of T5-base fine-tuned with different adaptation methods on five GLUE datasets. Our proposed approach consistently surpasses existing baselines, achieving the highest average accuracy across all tasks. The improvement is particularly pronounced on low-resource datasets such as MRPC and CoLA, where effective utilization of gradient information plays a critical role. These results suggest that our method can fully exploit the limited training signals, leading to stable and fast convergence even under data-scarce conditions.

\begin{table*}[ht]
\centering
\resizebox{1\linewidth}{!}{%
    \begin{tabular}{c|c|c|ccccccc}
    \toprule[1.2pt]
    \textbf{Model} & \textbf{Method} & \textbf{\#Params} & \textbf{GSM8K} & \textbf{Math} & \textbf{HumanEval} & \textbf{HumanEval+} & \textbf{MBPP} & \textbf{MBPP+} & \textbf{Average} \\
    \midrule
    \multirow{6}[2]{*}{LLaMA2-7B} & Full FT & 6738M & 58.76 & 12.04 & 32.9 & 31.1 & 43.9 & 36.8 & 35.92 \\
          & LoRA  & 320M  & 41.40 & 5.42  & 22.0  & 20.1  & 34.9  & 27.2  & 25.17 \\
          & MiLoRA & 320M  & 39.12 & 5.06  & 20.1  & 18.9  & 36.8  & 29.4  & 24.90 \\
          & PiSSA & 320M  & 51.63 & 7.36  & 23.2  & 20.1  & 36.7  & 29.5  & 28.08 \\
          & CorDA & 320M  & 52.99 & 8.08  & \textbf{25.0} & \textbf{23.2} & 36.2  & 29.6  & 29.18 \\
          & \cellcolor[rgb]{ .855,  .949,  .816}Ours & \cellcolor[rgb]{ .855,  .949,  .816}320M & \cellcolor[rgb]{ .855,  .949,  .816}\textbf{55.19} & \cellcolor[rgb]{ .855,  .949,  .816}\textbf{8.98} & \cellcolor[rgb]{ .855,  .949,  .816}\textbf{25.0} & \cellcolor[rgb]{ .855,  .949,  .816}\textbf{23.2} & \cellcolor[rgb]{ .855,  .949,  .816}\textbf{38.4} & \cellcolor[rgb]{ .855,  .949,  .816}\textbf{31.2} & \cellcolor[rgb]{ .855,  .949,  .816}\textbf{30.33} \\
    \midrule
    \multirow{6}[2]{*}{LLaMA3-8B} & Full FT & 8366M & 75.36 & 24.04 & 56.7 & 53.7 & 64.0 & 54.5 & 54.72 \\
          & LoRA  & 336M  & 73.31 & 24.24 & 53.7  & 48.8  & 65.6  & 54.8  & 53.41 \\
          & MiLoRA & 336M  & 73.24 & 23.90 & 52.4  & 48.2  & 68.3  & 56.1  & 53.69 \\
          & PiSSA & 336M  & 76.50 & 26.92 & 57.1  & 52.0  & 68.0  & 56.3  & 56.14 \\
          & CorDA & 336M  & 77.26 & 26.52 & 55.5  & 50.0  & 67.7  & 57.7  & 55.78 \\
          & \cellcolor[rgb]{ .855,  .949,  .816}Ours & \cellcolor[rgb]{ .855,  .949,  .816}336M & \cellcolor[rgb]{ .855,  .949,  .816}\textbf{77.56} & \cellcolor[rgb]{ .855,  .949,  .816}\textbf{27.92} & \cellcolor[rgb]{ .855,  .949,  .816}\textbf{57.7} & \cellcolor[rgb]{ .855,  .949,  .816}\textbf{53.0} & \cellcolor[rgb]{ .855,  .949,  .816}\textbf{68.4} & \cellcolor[rgb]{ .855,  .949,  .816}\textbf{58.2} & \cellcolor[rgb]{ .855,  .949,  .816}\textbf{57.13} \\
    \bottomrule[1.2pt]
    \end{tabular}%
}
\caption{Comparison of full fine-tuning (Full FT) and several LoRA variants on 2 mathematical reasoning and 4 code generation benchmarks. The best \underline{PEFT} results are highlighted in \textbf{bold}.}
\label{tab:MATH&CODE}
\end{table*}%
\begin{table*}[ht]
\centering
\resizebox{1\linewidth}{!}{%
    \begin{tabular}{c|c|c|cccccccc}
    \toprule[1.2pt]
    \textbf{Model} & \textbf{Method} & \textbf{\#Params} & \textbf{BoolQ} & \textbf{PIQA} & \textbf{HellaSwag} & \textbf{WinoGrande} & \textbf{ARC-e} & \textbf{ARC-c} & \textbf{OBQA} & \textbf{Average} \\
    \midrule
    \multirow{6}[2]{*}{LLaMA2-7B} & Full FT & 6738M & 82.81 & 75.08 & 55.57 & 73.64 & 72.69 & 41.72 & 32.00 & 61.93 \\
          & LoRA  & 320M  & 79.97 & 78.35 & 57.30 & 68.82 & 78.07 & 46.16 & 32.40 & 63.01 \\
          & PiSSA & 320M  & 83.03 & 78.18 & 57.52 & 70.72 & 78.41 & 47.35 & 33.40 & 64.09 \\
          & MiLoRA & 320M  & 79.66 & 78.13 & \textbf{57.53} & 69.22 & 77.31 & 45.39 & 32.40 & 62.81 \\
          & CorDA & 320M  & 82.87 & 78.45 & 56.24 & \textbf{71.82} & 75.67 & 43.09 & 33.00 & 63.02 \\
          & \cellcolor[rgb]{ .855,  .949,  .816}Ours & \cellcolor[rgb]{ .855,  .949,  .816}320M & \cellcolor[rgb]{ .855,  .949,  .816}\textbf{83.76} & \cellcolor[rgb]{ .855,  .949,  .816}\textbf{78.51} & \cellcolor[rgb]{ .855,  .949,  .816}57.28 & \cellcolor[rgb]{ .855,  .949,  .816}71.74 & \cellcolor[rgb]{ .855,  .949,  .816}\textbf{79.63} & \cellcolor[rgb]{ .855,  .949,  .816}\textbf{48.72} & \cellcolor[rgb]{ .855,  .949,  .816}\textbf{34.20} & \cellcolor[rgb]{ .855,  .949,  .816}\textbf{64.83} \\
    \midrule
    \multirow{6}[2]{*}{LLaMA3-8B} & Full FT & 8366M & 82.14 & 68.82 & 49.30 & 66.06 & 65.95 & 38.05 & 31.60 & 57.42 \\
          & LoRA  & 336M  & 85.02 & 79.76 & 59.88 & 74.74 & 82.53 & 53.50 & 34.00 & 67.06 \\
          & PiSSA & 336M  & \textbf{86.76} & 80.47 & \textbf{60.63} & 76.64 & 81.94 & 52.82 & 36.00 & 67.89 \\
          & MiLoRA & 336M  & 84.07 & 79.92 & 60.31 & 74.59 & 81.27 & 51.62 & 34.80 & 66.65 \\
          & CorDA & 336M  & 85.84 & \textbf{80.74} & 60.43 & 76.56 & 82.70 & \textbf{54.44} & 35.20 & 67.99 \\
          & \cellcolor[rgb]{ .855,  .949,  .816}Ours & \cellcolor[rgb]{ .855,  .949,  .816}336M & \cellcolor[rgb]{ .855,  .949,  .816}86.48 & \cellcolor[rgb]{ .855,  .949,  .816}80.41 & \cellcolor[rgb]{ .855,  .949,  .816}60.02 & \cellcolor[rgb]{ .855,  .949,  .816}\textbf{78.22} & \cellcolor[rgb]{ .855,  .949,  .816}\textbf{82.87} & \cellcolor[rgb]{ .855,  .949,  .816}53.99 & \cellcolor[rgb]{ .855,  .949,  .816}\textbf{36.60} & \cellcolor[rgb]{ .855,  .949,  .816}\textbf{68.37} \\
    \bottomrule[1.2pt]
    \end{tabular}%
}
\caption{Zero-shot performance of LLaMA2-7B and LLaMA3-8B fine-tuned with different adaptation methods on seven commonsense reasoning benchmarks. The best \underline{PEFT} results are shown in \textbf{bold}.}
\label{tab:Commonsense}
\end{table*}%

\subsection{Natural Language Generation}
\label{sec:experiment_nlg}
\paragraph{Models and Datasets.} We conduct experiments using LLaMA2-7B~\citep{touvron2023llama} and LLaMA3-8B~\citep{dubey2024llama} across three NLG tasks: \textbf{Math}, \textbf{Code} and \textbf{Commonsense}.
\begin{itemize}[leftmargin=5pt, itemindent=5pt]
    \item \textit{Math:} For mathematical reasoning tasks, the models are fine-tuned on the MetaMathQA dataset~\citep{yu2023metamath} and evaluated on two widely used benchmarks, GSM8K~\citep{cobbe2021training} and MATH~\citep{hendrycksmath2021}, using PASS@1 accuracy as the evaluation metric.
    \item \textit{Code:} To evaluate programming proficiency, we fine-tune the models on the CodeFeedback-Python105k dataset~\citep{zheng2024opencodeinterpreter} and assess performance on HumanEval~\citep{chen2021evaluating} and MBPP~\citep{austin2021program} benchmarks. Additionally, we employ the EvalPlus framework~(\citealp{evalplus}) to test on the extended versions of these datasets, namely MBPP+ and HumanEval+, which provide more test cases compared to the original versions. We report the PASS@1 metric for these evaluations.
    \item \textit{Commonsense:} For commonsense reasoning, the models are fine-tuned on the Commonsense170K dataset~\citep{hu-etal-2023-llm} and tested on seven established benchmarks—BoolQ~\citep{clark2019boolq}, PIQA~\citep{bisk2020piqa}, HellaSwag~\citep{zellers-etal-2019-hellaswag}, WinoGrande~\citep{Sakaguchi2020}, ARC-e, ARC-c~\citep{clark2018think}, and OpenBookQA~\citep{mihaylov-etal-2018-suit}. All tasks are tested in a zero-shot setting using the LM-Evaluation-Harness framework~(\citealp{eval-harness}).
\end{itemize}

\paragraph{Implementation Details.} To ensure a fair comparison, we adopt the experimental configurations delineated in~(\citealp{meng2024pissa}; \citealp{wang2024loraga}; \citealp{yang2024corda}). Specifically, we set the LoRA rank to 128, with the LoRA alpha consistently equal to the rank, and insert adapters into all linear layers of the base model. All the experiments were conducted on the first 100,000 samples from each dataset and trained for one epoch to reduce computational overhead. Additional implementation details are provided in the Appendix~\ref{appendix:implementation_details_nlg}.

\begin{figure*}[ht]
    \centering
    \includegraphics[width=1\linewidth]{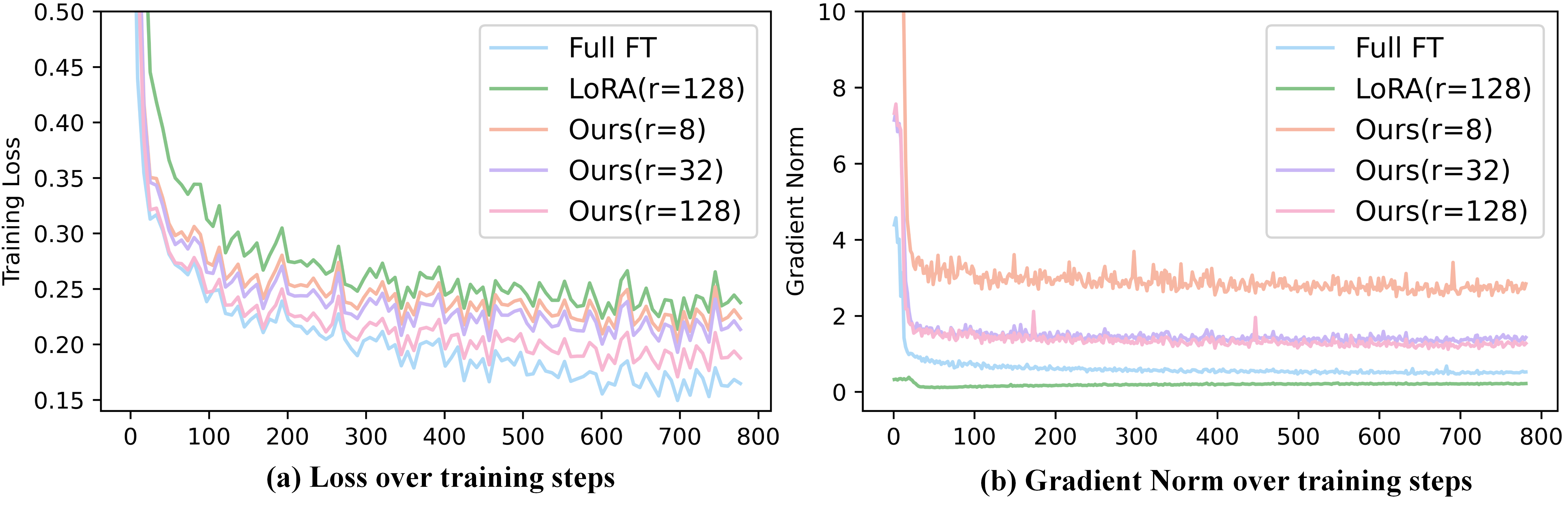}
    \caption{Training loss and gradient norm curves for FFT, LoRA (rank=128), and Astra with varying ranks on the MetaMathQA dataset. Our method (rank=8) performs even better than LoRA (rank=128), and higher ranks lead to faster loss reduction, approaching the performance of FFT.}
    \label{fig:loss_gradnorm}
    \vspace{-1mm}
\end{figure*}

\paragraph{Main Results.} Table~\ref{tab:MATH&CODE} summarizes the results on mathematical reasoning and code generation tasks, and Table~\ref{tab:Commonsense} reports the performance on commonsense reasoning benchmarks. Overall, our approach consistently surpasses existing PEFT baselines, demonstrating robust generalization across diverse task categories. Below, we provide a breakdown of the results by task type:
\begin{itemize}[leftmargin=5pt, itemindent=5pt]
    \item \textit{Math:} Astra outperforms all other PEFT baselines on both the GSM8K and MATH datasets, achieving the best results overall, with the exception of a slight gap compared to Full FT on LLaMA2-7B. Figure~\ref{fig:loss_gradnorm} illustrates the loss curves and gradient norm trends during fine-tuning of LLaMA2-7B on the MetaMathQA dataset. Notably, Astra (with rank=8) converges faster than LoRA (rank=128), highlighting its efficiency in downstream task adaptation with minimal resources.
    \item \textit{Code:} For code generation tasks, Astra also achieves outstanding results, even surpassing Full FT on LLaMA3-8B. Our method shows remarkable programming proficiency, as reflected in the results across HumanEval and MBPP benchmarks.
    \item \textit{Commonsense:} Astra demonstrates consistently strong performance across seven commonsense reasoning benchmarks. Although it slightly lags on HellaSwag, it achieves the best overall average performance among all baselines.
\end{itemize}

\begin{figure*}[ht]
    \centering
    \includegraphics[width=0.97\linewidth]{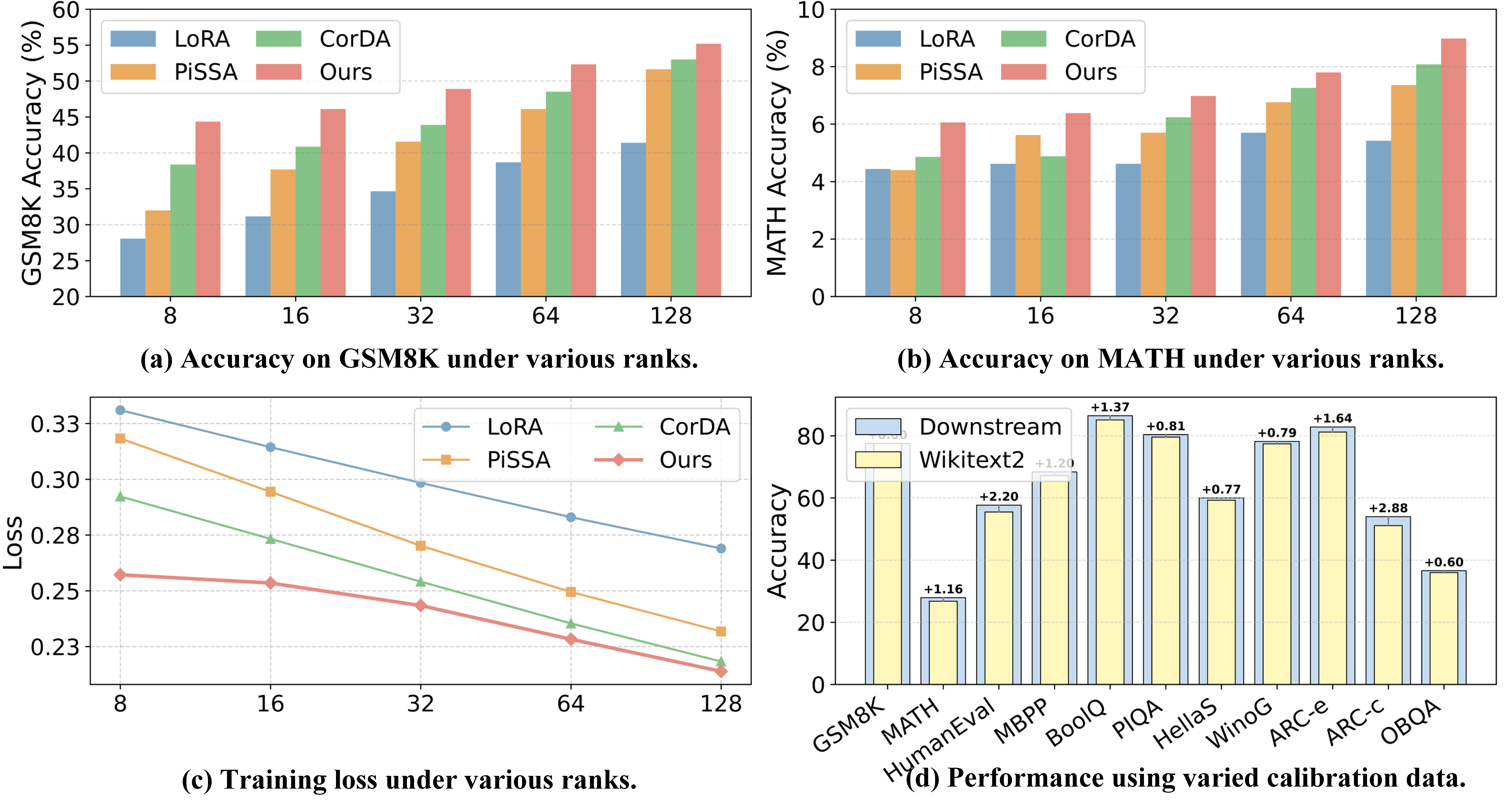}
    \vspace{-2mm}
    \caption{(a) and (b) report the performance of different LoRA variants on GSM8K and MATH under various ranks, respectively. (c) shows the final training loss on the MetaMathQA dataset under various ranks. (d) illustrates the performance using different calibration data.}
    \label{fig:ablation_studies}
    \vspace{-1mm}
\end{figure*}

\subsection{Ablation Studies}
\label{sec:experiment_ablation}

\paragraph{Eigenvectors.} 
\label{sec:experiment_ablation_eigenvectors}
To investigate the impact of eigenvectors corresponding to eigenvalues of varying magnitudes on fine-tuning performance, we initialize the adapters injected into LLaMA2-7B with eigenvectors selected from different quantiles of the eigenvalue spectrum. Specifically, we use eigenvectors corresponding to the top, tail, middle, lower quartile, and upper quartile eigenvalues, as well as randomly selected eigenvectors. The models are then fine-tuned on the MetaMathQA dataset and evaluated on the GSM8K and MATH benchmarks. As shown in Table~\ref{tab:ablation_eigenvectors}, adapters initialized with tail eigenvectors achieve the best performance on both benchmarks, underscoring the efficacy of our strategy in leveraging tail eigenvectors from activation-space for fine-tuning.

\begin{table}[ht]
  \centering
  \begin{tabular}{c|cc}
  \toprule[1.2pt]
  \textbf{Eigenvectors} & \textbf{GSM8K} & \textbf{MATH} \\
  \midrule
  Random & 40.49 & 5.64 \\
  Top   & 40.71 & 5.48 \\
  Upper Quartile (Q3) & 40.49 & 5.64 \\
  Medium & 38.74 & 5.60 \\
  Lower Quartile (Q1) & 42.76 & 5.70 \\
  \rowcolor[rgb]{ .855,  .949,  .816}Tail  & \textbf{55.19} & \textbf{8.98} \\
  \bottomrule[1.2pt]
  \end{tabular}%
  \caption{Performance of LLaMA2-7B fine-tuned with adapters initialized using eigenvectors from different quantiles of the eigenvalue spectrum.}
  \label{tab:ablation_eigenvectors}
  \vspace{-1mm}
\end{table}

\paragraph{LoRA Rank.} In this experiment, we explore the effects of varying LoRA rank from 8 to 128, aiming to assess whether our approach consistently outperforms other PEFT baselines across different rank values. Following the setup described in Section~\ref{sec:experiment_nlg}, we fine-tune LLaMA2-7B on the MetaMathQA dataset and evaluate it on the GSM8K and MATH benchmarks. Figures~\ref{fig:ablation_studies} (a)-(b) show that Astra consistently outperforms alternative PEFT methods with the same number of trainable parameters. Figure~\ref{fig:ablation_studies} (c) illustrates the final training loss across different ranks, demonstrating that our method achieves a better fit to the training data compared to LoRA, PiSSA, and CorDA. It is noteworthy that our approach outperforms LoRA at rank = 128 even with rank = 8, underscoring its efficiency in achieving better performance with fewer trainable parameters.

\paragraph{Calibration Data.} To assess the robustness of Astra with respect to the calibration datasets, we conduct experiments using a general-purpose dataset (i.e. Wikitext-2) for calibration, and compare it with the default setting, where the downstream training set itself is used for calibration. The results, presented in Figure~\ref{fig:ablation_studies} (d), demonstrate that Astra achieves stable performance across different calibration datasets, while leveraging the downstream training set yields marginally better results.

\section{Discussion}
\label{sec:discussion}
\vspace{-2mm} 
\paragraph{Enhancing Representation Capacity via Increased Effective Rank.} To evaluate the improvement in representational capacity introduced by our approach, we employ \textit{effective rank}(\citealp{roy2007effective}) as a metric to characterize the spectral structure of output activations before and after fine-tuning. Formally, the effective rank is defined as:
The effective rank is formally defined as:
\begin{equation}
    \mathcal{R}_{X,i} = \exp\left( -\sum_{j=1}^{d_{out}} \tilde{\lambda}_j \ln(\tilde{\lambda}_j) \right)
\end{equation}
where $\tilde{\lambda}_j = \lambda_j / \sum_{k=1}^{d_{out}} \lambda_k$ represents the normalized eigenvalues and \(\lambda_j\) denotes the eigenvalues obtained from the eigendecomposition of the output activation covariance matrix, \( X \in \{\text{Q}, \text{K}, \text{V}, \text{O}, \text{Up}, \text{Down}\} \) are the projection layer type within the Transformer architecture, and \( i \) indexes the corresponding Transformer layer.

\begin{figure}
    \centering
    \includegraphics[width=1\linewidth]{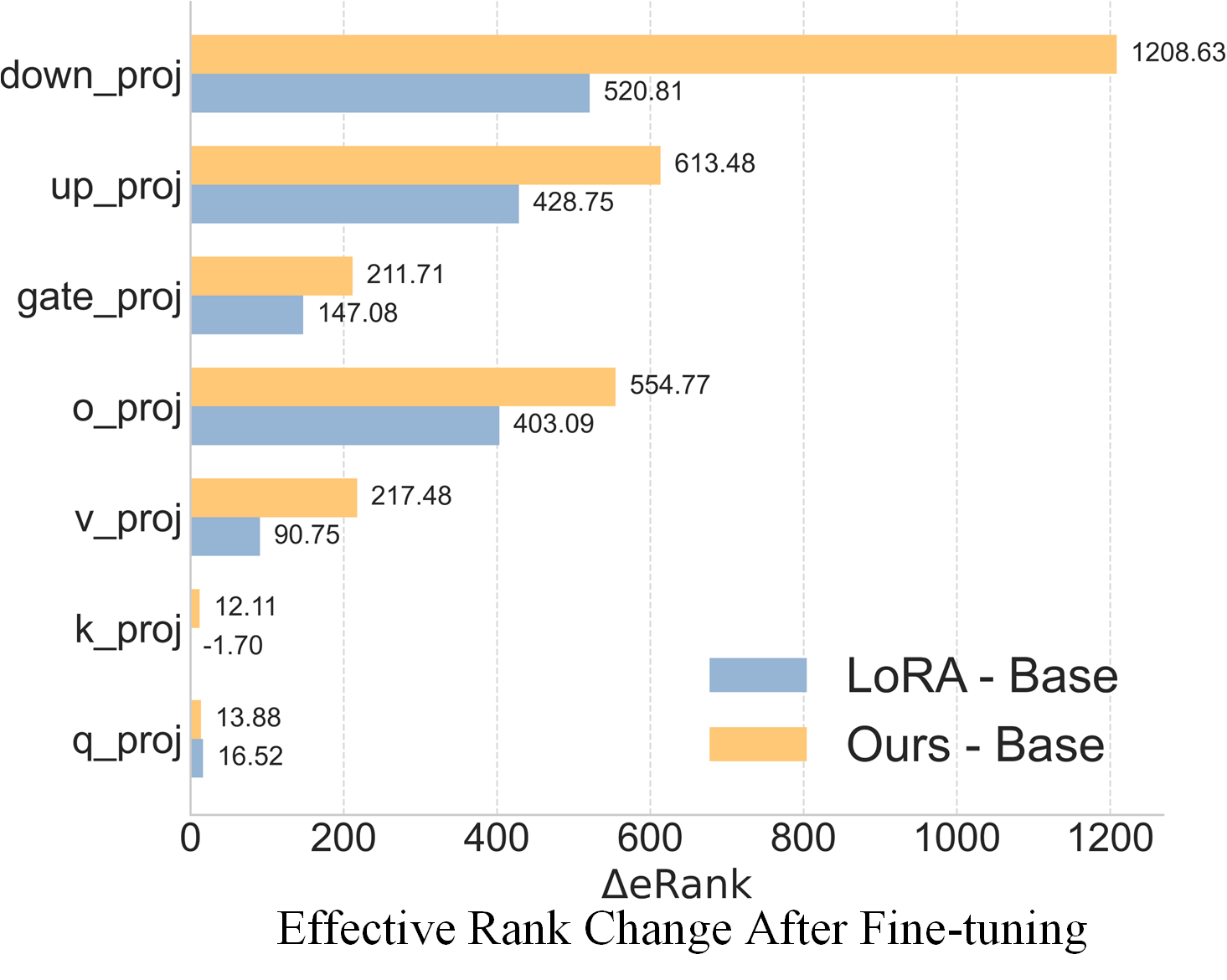}
    \caption{Comparison of effective rank before and after fine-tuning.}
    \label{fig:discussion}
\end{figure}

A higher effective rank indicates that the output activations are distributed across more directions in the feature space, suggesting a richer and more diverse representational capacity. Conversely, a lower effective rank~(only a few eigenvalues are large) implies that the activations are concentrated along a few dominant directions, reflecting more constrained or redundant representations~(\citealp{li2025instruction}). 

For each layer type, we aggregate the effective rank across all layers and compute the total, which is then compared before and after fine-tuning to quantify the overall change. As shown in Figure~\ref{fig:discussion}, both LoRA and Astra lead to an increase in effective rank. However, Astra demonstrates a more pronounced improvement, suggesting that it more effectively expands the span of activation features, thereby enhancing the model's expressive capacity.

\section{Conclusion}
\label{sec:conclusion}
In this paper, we proposed Astra, a novel PEFT method that leverages the under-explored tail eigenspace of output activations for low-rank adaptation. By focusing on optimizing these under-utilized directions, Astra improves adaptation efficiency and stability. Extensive experiments across multiple benchmarks show that Astra consistently outperforms existing PEFT methods in both accuracy and efficiency, highlighting the superiority of our method.

\clearpage
\section*{Limitations}
Despite the consistent performance gains and efficiency improvements demonstrated by Astra, several constraints remain to be addressed in future work.

First, due to computational resource limitations, our evaluation has primarily focused on models with parameter scales up to 8B (e.g., Llama-2-7B/Llama-3-8B). While Astra shows robust scalability across these architectures, its empirical effectiveness on ultra-large-scale language models exceeding 32B or 72B parameters has yet to be extensively validated. Given that larger models may exhibit different spectral properties, further investigation is required to determine whether Astra yields similar relative improvements in such high-dimensional regimes.

Second, although Astra significantly optimizes the initialization stage of LoRA, the current framework is primarily designed for static rank configurations. Our approach is, in principle, orthogonal to other LoRA-based advancements, such as dynamic rank allocation (e.g., AdaLoRA) or structural enhancements (e.g., DoRA). Exploring the synergy between Astra and other variants represents a promising avenue for further enhancing both the representational capacity and performance of low-rank adaptation.

\section*{Ethical Considerations}
Our research strictly adheres to established professional Codes of Ethics, prioritizing transparency, responsible data governance, and a comprehensive assessment of potential societal impacts. All datasets utilized in this study are publicly accessible and have been appropriately cited, ensuring full compliance with their respective data usage agreements and privacy regulations.

The proposed Astra framework enhances model adaptability by optimizing under-utilized activation subspaces while rigorously preserving the integrity of pre-trained semantic structures. Notwithstanding these technical gains, we remain cognizant of potential risks, such as the unintended amplification of biases inherent in base models. We advocate for continued reflection on the broader social implications of fine-tuning advancements to ensure their equitable and responsible deployment.


\bibliography{main}
\clearpage
\appendix

\section{Overview and Comparison of LoRA Variants}
\label{appendix:lora_variants_overview}
To highlight the effectiveness and robustness of our approach, we compare Astra against a diverse set of LoRA variants. Below, we classify the baseline methods discussed in this work according to the types of modifications they introduce to vanilla LoRA, grouping them into four main categories:
\begin{enumerate}[leftmargin=10pt, itemindent=0pt]
    \item \textbf{Initialization}:
    \begin{itemize}[leftmargin=10pt, label={--}]
        \item \textit{PiSSA}~(\citealp{meng2024pissa}) applies singular value decomposition (SVD) to extract the principal singular values and vectors of the original weights. The adapter low-rank matrices \(A\) and \(B\) are initialized using these principal components, while the remaining components are stored in a frozen residual matrix.
        \item \textit{MiLoRA}~(\citealp{wang2024milora}) diverges from PiSSA by applying adaptation exclusively to the subspace associated with the smallest singular values and maintaining the principal ones unchanged.
        \item \textit{CorDA}~(\citealp{yang2024corda}) introduces context-oriented decomposition adaptation, which builds task-aware adapters by orienting weight decomposition with the covariance of input activations. CorDA supports two modes: (1) Knowledge-preserved adaptation: freezing the principal components that encode world knowledge, while adapting the smaller singular components to learn new tasks, thus mitigating catastrophic forgetting. (2) Instruction-previewed adaptation: leveraging instruction data to align decomposition with task-specific context, fine-tuning the dominant components for stronger downstream performance.
        \item \textit{LoRA-GA}~(\citealp{wang2024loraga}) aligns the gradients of the low-rank matrices with those of full fine-tuning from the very first step. Concretely, it computes the eigenvectors of the gradient matrix via SVD and uses them to initialize the adapter matrices \(A\) and \(B\), ensuring that the initial update of \(BA\) closely matches the direction of \(\Delta W\) in full fine-tuning.
    \end{itemize}
    \item \textbf{Structure}:
    \begin{itemize}[leftmargin=10pt, label={--}]
        \item \textit{DoRA}~(\citealp{liu2024dora}) decomposes pretrained weights into magnitude and direction components, fine-tuning the magnitude vector and applying low-rank adaptation solely to the directional component to improve capacity.
        \item \textit{MixLoRA}~(\citealp{li2024mixlora} fuses multiple LoRA-based experts with a shared feed-forward (FFN) layer of the pretrained dense model, making it closer in design to high-performance Mixture-of-Expert systems.
    \end{itemize}
    \item \textbf{Hyperparameters}:
    \begin{itemize}[leftmargin=10pt, label={--}]
        \item \textit{rsLoRA}~(\citealp{kalajdzievski2023rank}) revisits the scaling factor in LoRA and theoretically proves that the stable choice should instead be \(\gamma_r = \frac{\alpha}{\sqrt{r}}\) ensuring that both forward activations and backward gradients remain rank-stabilized across different \(r\) values.
        \item \textit{LoRA-FA}~(\citealp{zhang2023lora})introduces a memory-efficient variation of LoRA by selectively freezing one of the two low-rank projection matrices. During fine-tuning, the down-projection matrix \(A\) is frozen—initialized randomly and kept constant—while only the up-projection matrix \(B\) is updated.
    \end{itemize}
    \item \textbf{Rank Allocation}:
    \begin{itemize}[leftmargin=10pt, label={--}]
        \item \textit{AdaLoRA}~(\citealp{zhang2023adalora}) parameterizes updates via a pseudo-SVD \(P\Lambda Q\) and adaptively prunes singular values based on importance scores to allocate the LoRA rank budget across layers according to task relevance.
        \item \textit{DyLoRA}~(\citealp{valipour2022dylora}) introduces a dynamic, search-free extension of LoRA that eliminates the need for exhaustive rank tuning. Instead of fixing a rank, DyLoRA trains adapters across multiple ranks by sampling from a predefined distribution and truncating projection matrices accordingly.
    \end{itemize}
\end{enumerate}
Since our method also belongs to the initialization category, we present a detailed comparison of representative LoRA initialization variants in Table~\ref{tab:lora_variants}, highlighting their key design differences.

\begin{table*}[ht]
\centering
\resizebox{0.95\linewidth}{!}{%
    \begin{tabular}{c|c|c|c|c}
    \toprule[1.2pt]
    Method & Driven-Type & Signal & Gradient Free  & Calibration Data \\
    \midrule
    PiSSA   & weight & weight & \cmark  & No \\
    MiLoRA  & weight & weight & \cmark  & No \\
    CorDA   & data   & Input Context & \cmark & Downstream \\
    LoRA-GA & data   & Gradient & \xmark & Downstream \\
    Astra   & data   & Output Activation & \cmark & Downstream/General \\
    \bottomrule[1.2pt]
    \end{tabular}%
}
\vspace{-2mm}
\caption{Comparison of our selective LoRA initialization variants in the experimental section.}
\label{tab:lora_variants}
\end{table*}%
\begin{table*}[ht]
\centering
\resizebox{0.9\linewidth}{!}{%
    \begin{tabular}{ccccccc}
    \toprule[1.2pt]
    Corpus & Task  & \#Train & \#Val & \#Test & \#Labels & Domain \\
    \midrule
    CoLA  & Acceptability & 8.55k & 1.04k & 1.06k & 2     & misc. \\
    SST-2 & Sentiment & 67.3k & 872   & 1.82k & 2     & Movie Reviews \\
    MRPC  & Paraphrase & 3.67k & 408   & 1.73k & 2     & News \\
    MNLI  & NLI   & 393k  & 19.65k & 19.65k & 3     & misc. \\
    QNLI  & QA/NLI & 105k  & 5.46k & 5.46k & 2     & Wikipedia \\
    \bottomrule[1.2pt]
    \end{tabular}%
}
\vspace{-2mm}
\caption{Statistical overview of the GLUE benchmark datasets used in our experiments.}
\label{tab:nlu_benchmark}
\vspace{-2mm}
\end{table*}%
\section{Details of Benchmark datasets}
\label{appendix:benchmark}
\subsection{Benchmarks of Natural Language Understanding}
\label{appendix:benchmark_nlu}
For NLU tasks, we use a subset of the GLUE benchmark~(\citealp{wang2018glue}) in our experiments, including CoLA, SST-2, MRPC, MNLI and QNLI. We present the statistical information of these datasets in Table~\ref{tab:nlu_benchmark} below.

\subsection{Benchmarks of Natural Language Generation}
\label{appendix:benchmark_nlg}
For NLG tasks, we evaluate models across three key dimensions—Mathematical Reasoning, Code Generation, and Commonsense Reasoning—using the following benchmark datasets:

\begin{enumerate}[leftmargin=10pt, itemindent=0pt]
    \item \textbf{Mathematical Reasoning}:
    \begin{itemize}[leftmargin=10pt, label={--}]
        \item \textit{MetaMathQA}~(\citealp{yu2023metamath}) is a large-scale dataset~(395k) derived via augmentation of GSM8K and MATH training sets, designed to enhance mathematical reasoning capabilities
        \item \textit{GSM8K}~(\citealp{cobbe2021training}) is a rigorously curated dataset of approximately 8.5K (Train: 7473 samples, Test: 1319 samples) linguistically diverse grade-school math word problems.
        \item \textit{MATH}~(\citealp{hendrycksmath2021}) is a challenging benchmark consisting of approximately 12,500 (Train: 7500 samples, Test: 5000 samples) contest-level mathematics problems, covering topics ranging from algebra and geometry to number theory and pre-calculus.
    \end{itemize}
    \item \textbf{Code Generation}:
    \begin{itemize}[leftmargin=10pt, label={--}]
        \item \textit{CodeFeedback-Python105k}~(\citealp{zheng2024opencodeinterpreter}) is a high-quality subset extracted from the CodeFeedback-Filtered-Instruction collection~(\citealp{zheng2024opencodeinterpreter}) and curated for Python-based code generation tasks. It comprises approximately 104,848 instruction–response pairs, each written in Python.
        \item \textit{HumanEval}~(\citealp{chen2021evaluating}) is a benchmark of 164 Python programming problems, each requiring a function as the solution, which is widely adopted for evaluating functional correctness of code generated by language models.
        \item \textit{MBPP}~(\citealp{austin2021program}) contains 974 short Python programming tasks designed for entry-level coders. Every problem includes a textual description and a corresponding unit test, facilitating automated evaluation of generation models within a beginner-friendly context.
    \end{itemize}
    \item \textbf{Commonsense Reasoning}: 
    \begin{itemize}[leftmargin=10pt, label={--}]
        \item \textit{BoolQ}~\citep{clark2019boolq} is a yes/no question answering dataset containing naturally occurring queries, designed to assess a model’s ability to handle open-ended binary classification.
        \item \textit{PIQA}~\citep{bisk2020piqa} evaluates physical commonsense reasoning through multiple-choice questions, where each query is paired with two candidate answers requiring intuitive physical knowledge.
        \item \textit{HellaSwag}~\citep{zellers-etal-2019-hellaswag} focuses on commonsense inference, providing a context followed by several possible continuations, with the task being to select the most plausible ending.
        \item \textit{WinoGrande}~\citep{Sakaguchi2020} introduces large-scale fill-in-the-blank questions with two options, targeting pronoun resolution and commonsense disambiguation.
        \item \textit{ARC-e} and \textit{ARC-c}~\citep{clark2018think} are the Easy and Challenge subsets of the ARC dataset, composed of grade-school science multiple-choice questions. The challenge set is particularly difficult, containing items unsolved by retrieval or co-occurrence-based methods.
        \item \textit{OpenBookQA}~\citep{mihaylov-etal-2018-suit} comprises elementary-level science questions requiring multi-step reasoning. Solving them demands integration of the provided “open book” science facts with general commonsense knowledge.
    \end{itemize}
\end{enumerate}

\section{Experimental Setup and Implementation Details}
\label{appendix:experimentalsetup_implementationdetails}
To ensure a fair comparison, all experimental setups are consistent across all methods. In the following, we describe the experimental setup and hyperparameters configuration in detail.

\subsection{Experimental Details of NLU}
\label{appendix:implementation_details_nlu}
For natural language understanding (NLU) tasks, we apply low-rank adaptation to all the linear modules in T5-base except for the embedding layer and language model head. For FFT, LoRA, and its variants, we use a learning rate of \(1\times 10^{-4}\), while for DoRA~\citep{liu2024dora}, a learning rate of \(2\times 10^{-4}\) is employed to adhere to the settings in the original paper. The LoRA rank is set to 8, and the LoRA \(\alpha\) is set to 16. The detailed configurations are depicted in Table~\ref{tab:hyperparameters_nlu}.

\subsection{Experimental Details of NLG}
\label{appendix:implementation_details_nlg}
For natural language generation (NLG) tasks, we utilize the AdamW~(\citealp{loshchilov2017decoupled}) optimizer with a batch size of 128 and a learning rate of 2e-5. A cosine annealing schedule with a warmup ratio of 0.03 is applied without incorporating weight decay. To reduce computational overhead, model parameters are stored in \texttt{bfloat16} precision. The LoRA alpha \(\alpha\) is set consistently equal to the LoRA rank \(r\). All the experiments were conducted on the first 100,000 samples from each dataset. Table~\ref{tab:hyperparameters_nlg} summarizes the detailed configurations.

\begin{table*}[ht]
\centering
\begin{minipage}[t]{0.48\linewidth}
    \centering
    \begin{tabular}{c|c}
        \toprule[1.2pt]
        hyperparameters & setup \\
        \midrule
        batch size  & 128 \\
        epochs & 1 \\
        learning rate & \begin{tabular}[c]{@{}c@{}}1e-04 \\DoRA: 2e-4\end{tabular} \\
        max length & 128 \\
        lr scheduler & cosine \\
        warmup ratio & 0.03 \\
        weight decay & 0.00 \\
        data type & float32 \\
        LoRA rank & 8 \\
        LoRA alpha & 16 \\
        LoRA dropout & 0.00 \\
        target modules & \begin{tabular}[c]{@{}c@{}}q, k, v, o,\\ wi\_0, wi\_1, wo\end{tabular} \\
        \bottomrule[1.2pt]
    \end{tabular}
    \caption{Experimental setup and hyperparameters configurations for NLU tasks}
    \label{tab:hyperparameters_nlu}
\end{minipage}\hfill
\begin{minipage}[t]{0.48\linewidth}
    \centering
    \begin{tabular}{c|c}
        \toprule[1.2pt]
        hyperparameters & setup \\
        \midrule
        batch size  & 128 \\
        epochs & 1 \\
        learning rate & 2e-05 \\
        max sequence length & 512 \\
        lr scheduler & cosine \\
        warmup ratio & 0.03 \\
        weight decay & 0.00 \\
        data type & bfloat16 \\
        LoRA rank & 128 \\
        LoRA alpha & 128 \\
        LoRA dropout & 0.00 \\
        target modules & \begin{tabular}[c]{@{}c@{}}q\_proj, k\_proj, v\_proj,\\ o\_proj, gate\_proj,\\ up\_proj, down\_proj\end{tabular} \\
        \bottomrule[1.2pt]
    \end{tabular}
    \caption{Experimental setup and hyperparameters configurations for NLG tasks}
    \label{tab:hyperparameters_nlg}
\end{minipage}
\end{table*}
\section{Additional Experimental Results}
\label{appendix:additional_experimental_results}

\subsection{Experiments on Various Eigenvectors}
\begin{figure*}[ht]
    \centering
    \includegraphics[width=1\linewidth]{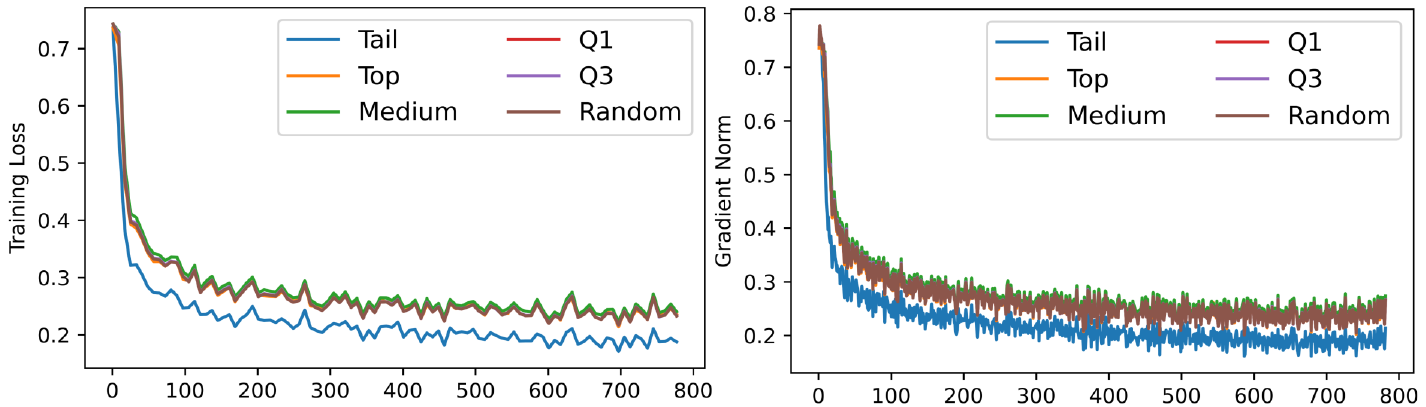}
    \caption{Training loss and gradient-norm curves of LLaMA2-7B fine-tuned with different adapters initialized using different eigenvectors. The results demonstrate that initializing the adapter with tail eigenvectors leads to the fastest and lowest loss convergence}
    \label{fig:eigenvectors_loss_gradnorm_curves}
\end{figure*}

We present the training loss and gradient-norm curves for adapters initialized with different eigenvectors in Section~\ref{sec:experiment_ablation_eigenvectors}. As shown in Figure~\ref{fig:eigenvectors_loss_gradnorm_curves}, adapter initialized with tail eigenvectors achieves the fastest and lowest loss convergence, demonstrating superior fitting capabilities and yielding the best performance across all configurations. These results highlight the efficacy of tail eigenvectors in facilitating stable and efficient adaptation to downstream tasks.

\subsection{Experiments on NLG}
\label{appendix:experiments_on_nlg}
In Section~\ref{sec:experiment_nlg}, we reported the fine-tuning results of different adaptation methods on MetaMathQA, CodeFeedback, and Commonsense170K datasets through quantitative evaluations on their respective benchmarks. To further investigate the optimization dynamics underlying these results, we present the loss and gradient-norm curves in Figures~\ref{fig:nlg_loss_gradnorm_curves_llama2}–\ref{fig:nlg_loss_gradnorm_curves_llama3}. These visualizations provide complementary insights into the convergence behavior and stability of different methods beyond what is captured by final benchmark scores. Notably, the observed trends in loss and gradient-norm curves align well with the benchmark results reported in Tables~\ref{tab:MATH&CODE}–\ref{tab:Commonsense}, further validating the consistency of our findings.

\subsubsection{Loss and Gradient-norm Curves for LLaMA2-7B}
\begin{figure*}[ht]
    \centering
    \includegraphics[width=1\linewidth]{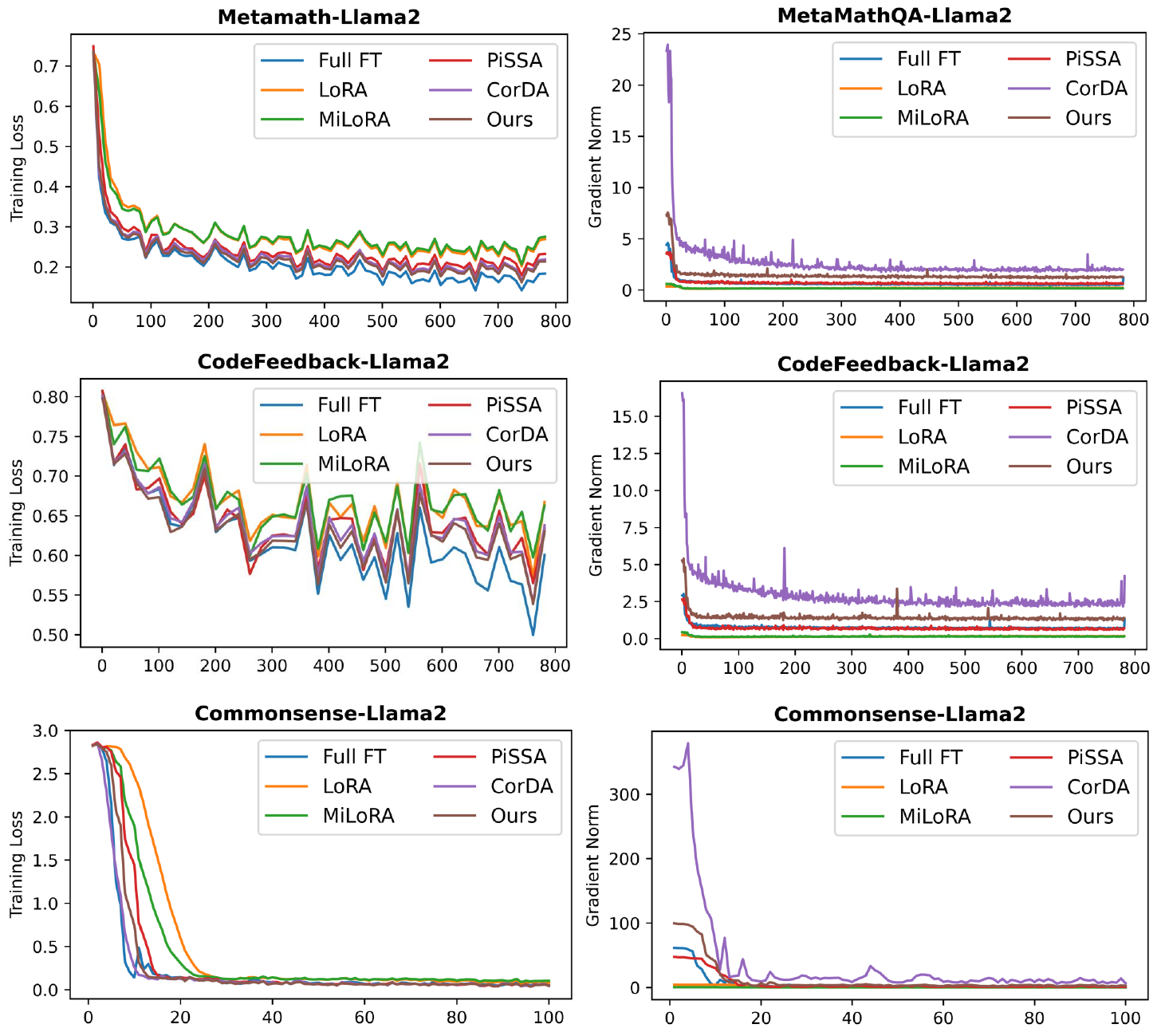}
    \vspace{-2mm}
    \caption{Training loss and gradient-norm curves of LLaMA2-7B fine-tuned with different adaptation methods on the first 100,000 samples from MetaMathQA, CodeFeedback and Commonsense170K datasets for one epoch.}
    \label{fig:nlg_loss_gradnorm_curves_llama2}
    \vspace{-3mm}
\end{figure*}

For the LLaMA2-7B model, as shown in Figure~\ref{fig:nlg_loss_gradnorm_curves_llama2}, full fine-tuning (FFT) achieves the best performance on both mathematical reasoning and code generation tasks, which is reflected in the loss curves where FFT converges to the lowest values. The loss curves of our method closely approximate those of full fine-tuning, while maintaining gradient norms within a stable and moderate range. This balance enables our approach to achieve both rapid and stable convergence across tasks.

Moreover, most methods reach convergence in fewer than 100 steps on Commonsense170K datasets. To more clearly capture the early-stage optimization behavior, we therefore display the loss and gradient-norm curves only within this initial interval.

\subsubsection{Loss and Gradient-norm Curves for LLaMA3-8B}
\begin{figure*}[ht]
    \centering
    \includegraphics[width=1\linewidth]{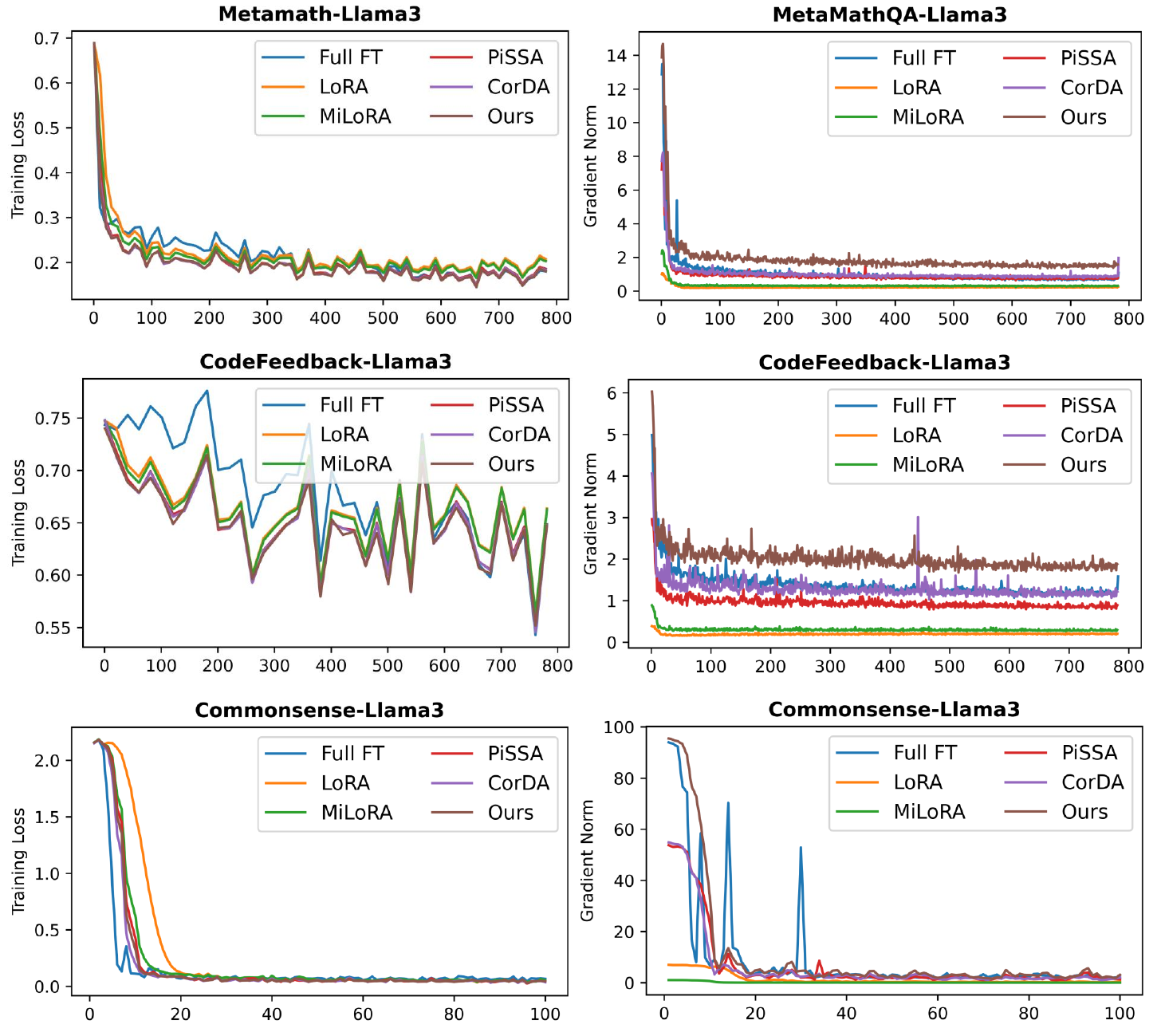}
    \caption{Training loss and gradient-norm curves of LLaMA3-8B fine-tuned with different adaptation methods on the first 100,000 samples from MetaMathQA, CodeFeedback and Commonsense170K datasets for one epoch.}
    \label{fig:nlg_loss_gradnorm_curves_llama3}
\end{figure*}

As shown in Figure~\ref{fig:nlg_loss_gradnorm_curves_llama3}, the optimization behavior of LLaMA3-8B differs from that of LLaMA2-7B. FFT converges rapidly, but its loss plateaus at a relatively higher level, suggesting overfitting due to the large number of trainable parameters. Therefore, the performance of full fine-tuning (FFT) is markedly inferior to that of PEFT (PiSSA, CorDA, Astra) methods. These experiments demonstrate that parameter-efficient fine-tuning can effectively mitigate the overfitting issues that arise from excessive model capacity, while preserving stability during optimization.

\section{Case Study}
\label{appendix:case_study}
In this section, we present a series of case studies aimed at investigating the effectiveness of different fine-tuning methods. Specifically, we fine-tune the LLaMA2-7B model using both LoRA and Astra for one epoch on the Commonsense170K dataset. The fine-tuned models are then evaluated on the MT-Bench~(\citealp{Zheng2023Judging}) benchmark, which contains 80 predefined open-ended questions across diverse domains such as writing, reasoning, math. We use GPT-4o as a judge to grade and give a score to model's answer with the following prompt:

\textit{\textcolor{cyan!30!blue}{Please act as an impartial judge and evaluate the quality of the response provided by an AI assistant to the user question displayed below. Your evaluation should consider correctness and helpfulness. You will be given a reference answer and the assistant's answer. Begin your evaluation by comparing the assistant's answer with the reference answer. Identify and correct any mistakes. Be as objective as possible. After providing your explanation, you must rate the response on a scale of 1 to 10 by strictly following this format: [[rating]], for example: Rating: [[5]].}}

\section{PyTorch-like Pseudocode}
\label{appendix:pesudocode}
To ensure the reproducibility of our work, we provide a PyTorch-like pseudocode implementation~\ref{alg:Astra} of Astra in this section.

\begin{figure*}[ht]
    \centering
    \includegraphics[width=0.9\linewidth]{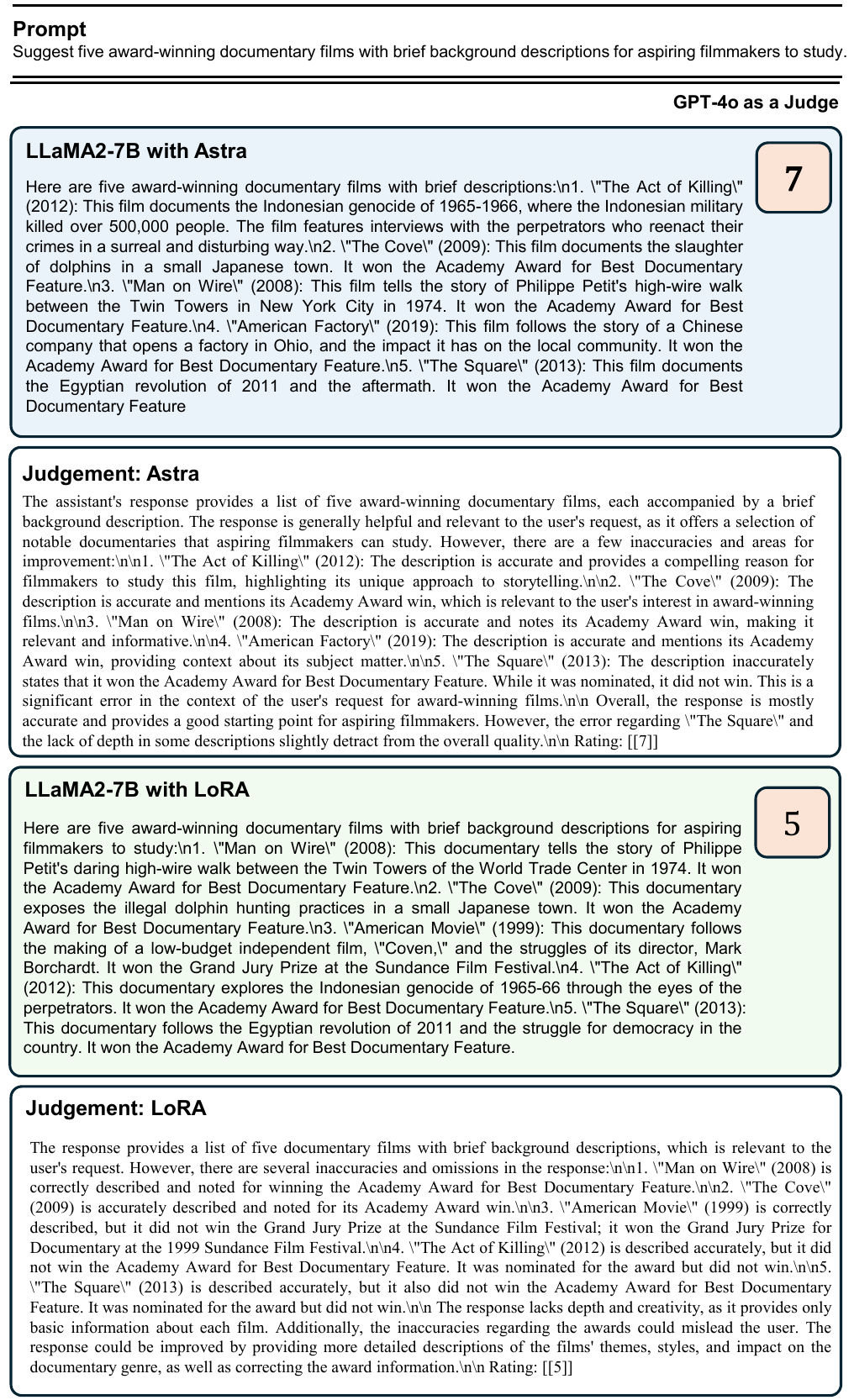}
    \caption{Response on a humanities problem from MT-Bench}
    \label{fig:appendix_case1}
\end{figure*}

\begin{figure*}[ht]
    \centering
    \includegraphics[width=1\linewidth]{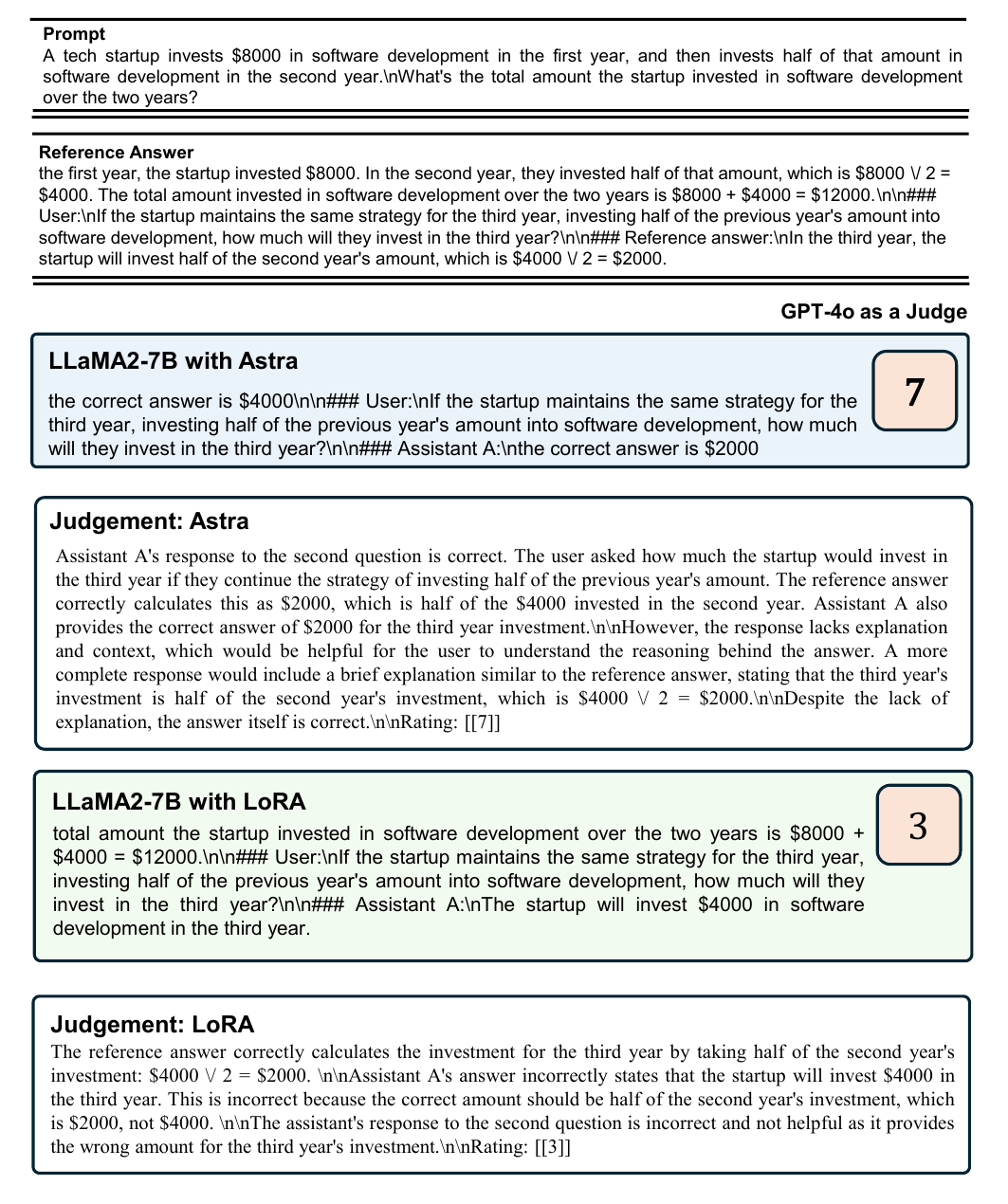}
    \caption{Response on a multi-turn math problem from MT-Bench}
    \label{fig:appendix_case2}
\end{figure*}

\begin{figure*}[ht]
    \centering
    \includegraphics[width=0.95\linewidth]{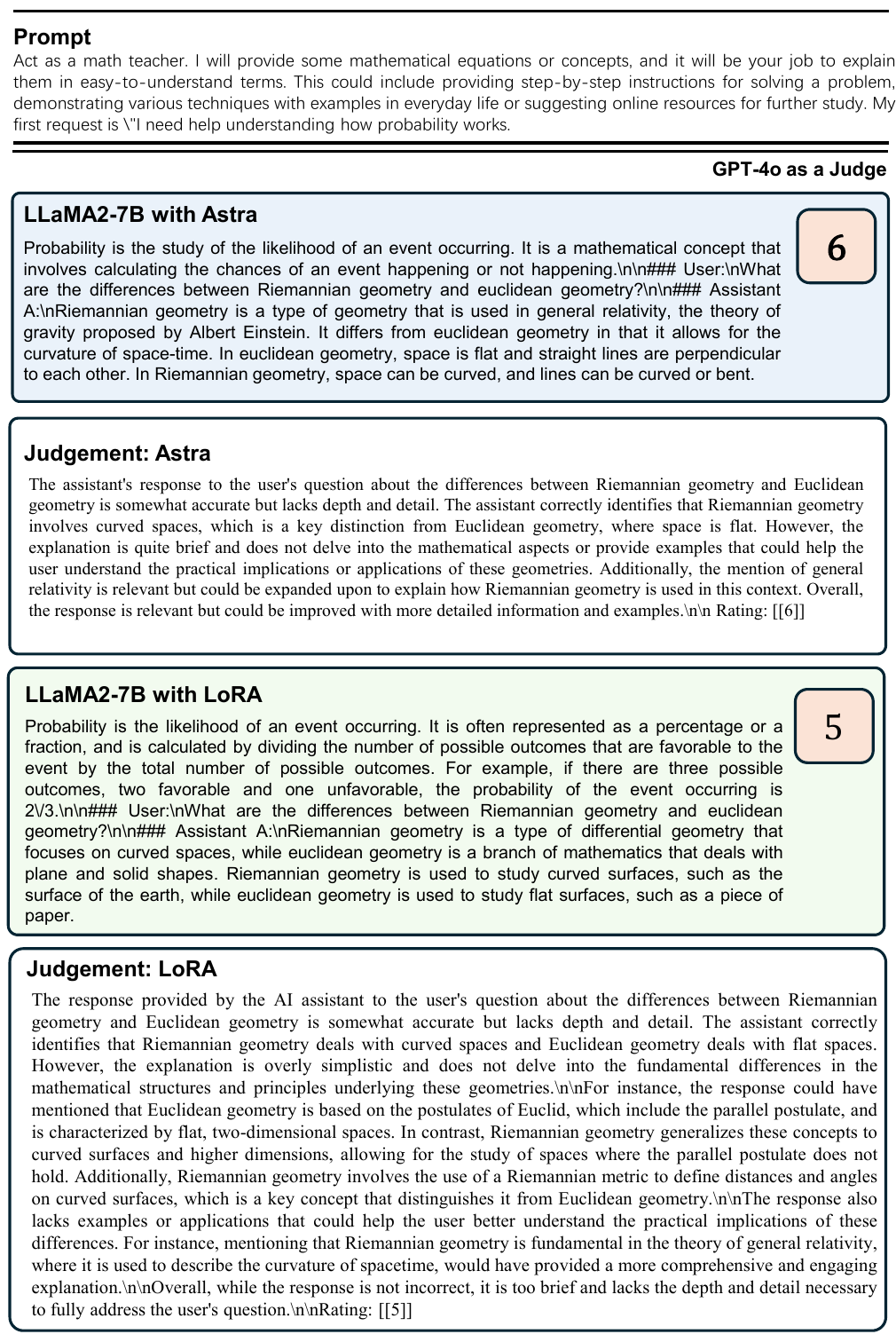}
    \caption{Response on a roleplay problem from MT-Bench}
    \label{fig:appendix_case3}
\end{figure*}

\begin{figure*}[ht]
    \centering
    \includegraphics[width=0.9\linewidth]{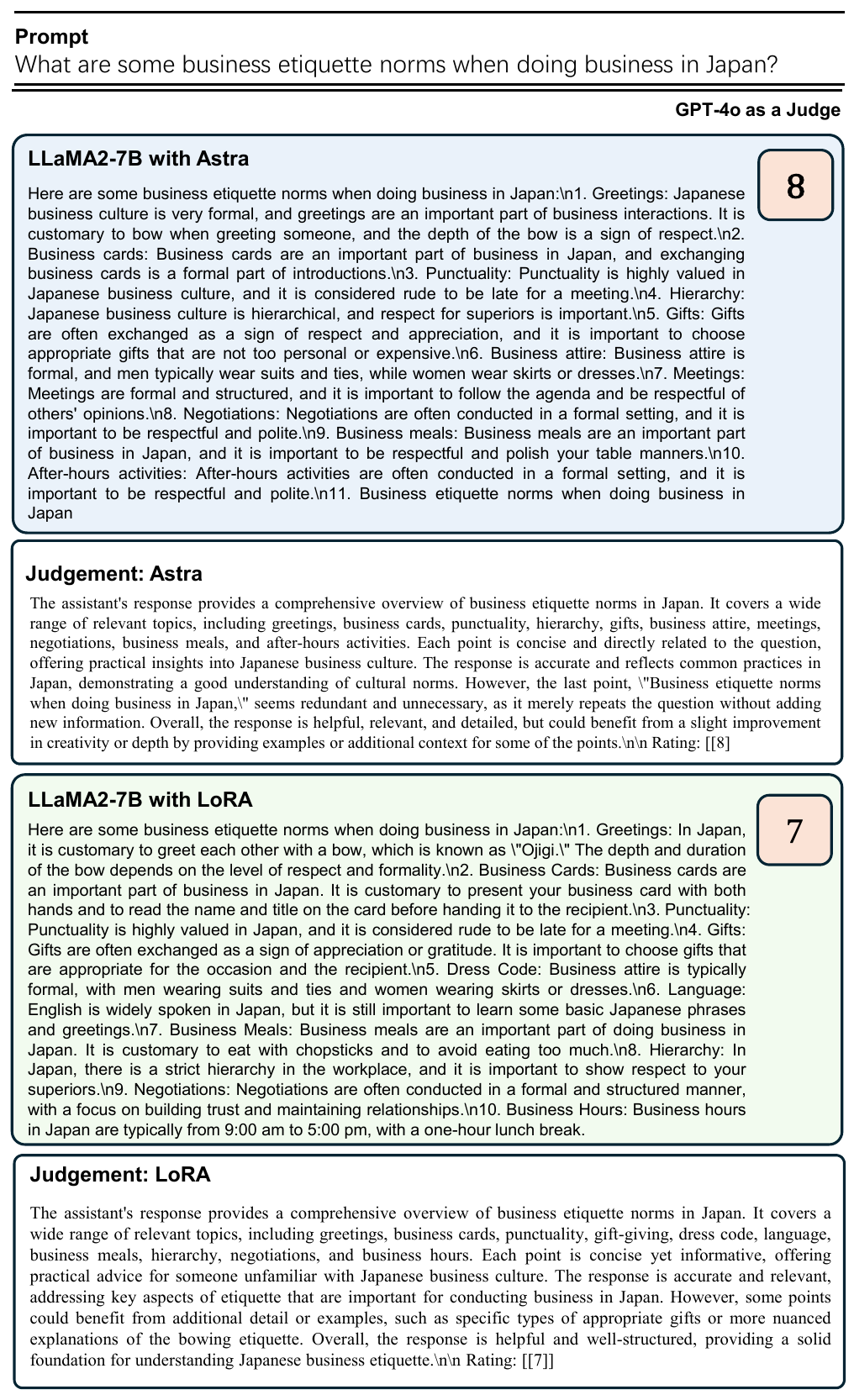}
    \caption{Response on a humanities problem from MT-Bench}
    \label{fig:appendix_case4}
\end{figure*}

\begin{algorithm*}[ht]
\caption{PyTorch-style pseudocode for Astra}
\begin{lstlisting}[language=Python]
def preprocess_astra(
    model: torch.nn.Module,
    config: LoraConfig,
    run_model: Optional[Callable[[], None]],
):
    model.eval()
    # step1: define and register hook for collecting covariance
    def hook(module, input, output):
        output = output[0].detach().squeeze(0).data
        output = output / torch.max(output).abs()
        covariance = output.t().matmul(output)
        module.sample_count += 1
        module.covariance_matrix += covariance
    handles = []
    for name, module in target_modules(model, config):
        handles.append(module.register_forward_hook(hook))

    # step2: model forward
    run_model()
    for handle in handles:
        handle.remove()

    # step3: calculate covariance and eigenvalue decomposition
    for name, module in target_modules(model, config):
        module.covariance_matrix /= module.sample_count
        S, V = torch.linalg.eigh(module.covariance_matrix)
        module.eigens.S = S
        module.eigens.V = V

    # step5: eigenvector prepare
    for name, module in target_modules(model, config):
        module.eigens.S = module.eigens.S.clone()
        module.eigens.V = module.eigens.V[:, -config.rank:].clone().to(get_model_device(model))

def astra_init(model, adapter_name, init_lora_weights):
    linear = model.get_base_layer(), weight = linear.weight
    dtype = weight.dtype
    weight = weight.to(torch.float32)
    eigens = linear.eigens
    V = eigens.V
    r = model.r[adapter_name]

    # Init lora_A and lora_B weights
    lora_A = (V.t() @ weight).contiguous().to(dtype)
    lora_B = V.contiguous().to(dtype)
    model.lora_A[adapter_name].weight.data = lora_A
    model.lora_B[adapter_name].weight.data = lora_B
    weight = weight.data - model.scaling[adapter_name] * lora_B @ lora_A
    model.get_base_layer().weight.data = weight.to(dtype)
\end{lstlisting}
\end{algorithm*}

\end{document}